\documentclass{article}

     \PassOptionsToPackage{square, numbers}{natbib}

\usepackage[preprint]{neurips_2022}

\usepackage[utf8]{inputenc} 
\usepackage[T1]{fontenc}    
\usepackage{hyperref}       
\usepackage{url}            
\usepackage{booktabs}       
\usepackage{amsfonts}       
\usepackage{nicefrac}       
\usepackage{microtype}      
\usepackage{xcolor}         
\usepackage{graphicx}
\usepackage{bbding}
\usepackage{amsmath}
\usepackage{wrapfig}

\title{PTCT: Patches with 3D-Temporal Convolutional Transformer Network for Precipitation Nowcasting}

\author{
  Ziao Yang \\
  School of Data and Computer Science\\
  Sun Yat-Sen University \\
  \texttt{yangzao@mail2.sysu.edu.cn} \\
   \And
   Xiangrui Yang \\
   School of Data and Computer Science\\
    Sun Yat-Sen University \\
   \texttt{yangxr9@mail2.sysu.edu.cn} \\
   \AND
   Qifeng Lin \thanks{Corresponding author} \\
   School of Data and Computer Science\\
  Sun Yat-Sen University \\
   \texttt{linqf6@mail2.sysu.edu.cn} \\
}

\begin{document}

\maketitle

\begin{abstract}

Precipitation nowcasting is to predict the future rainfall intensity over a short period of time, which mainly relies on the prediction of radar echo sequences. Though convolutional neural network (CNN) and recurrent neural network (RNN) are widely used to generate radar echo frames, they suffer from inductive bias (i.e., translation invariance and locality) and seriality, respectively. Recently, Transformer-based methods also gain much attention due to the great potential of Transformer structure, whereas short-term dependencies and autoregressive characteristic are ignored. In this paper, we propose a variant of Transformer named patches with 3D-temporal convolutional Transformer network (PTCT), where original frames are split into multiple patches to remove the constraint of inductive bias and 3D-temporal convolution is employed to capture short-term dependencies  efficiently. After training, the inference of PTCT is performed in an autoregressive way to ensure the quality of generated radar echo frames. To validate our algorithm, we conduct experiments on two radar echo dataset: Radar Echo Guangzhou and HKO-7. The experimental results show that PTCT achieves state-of-the-art (SOTA) performance compared with existing methods.
\end{abstract}

\section{Introduction}

Precipitation nowcasting is to predict the future rainfall intensity over a short period of time (e.g., 0$\sim$2 hours), which is crucial for numerous weather-dependent sectors to reduce losses caused by extreme weather, e.g., air traffic control, agriculture and tourist industry \cite{xingjian2015convolutional, wilson2010nowcasting}. Traditional methods are based on numerical weather prediction (NWP), which requires a complex and meticulous design of the physical model to represent the atmosphere state exactly \cite{lorenc1986analysis}. However, building such a model and solving large-scale system of equations are difficult and time-consuming, which motivates radar echo based approaches \cite{shi2017deep}. Generating consecutive and accurate radar echo frames is crucial for judging whether there exits rainfall or not, which needs to consider both temporal and spatial dependencies simultaneously. Temporal dependencies consist of short-term and long-term  information in the temporal dimension, while spatial dependencies contain short-range and long-range information in the spatial dimension.

 With impressive development of neural networks in various fields, how to apply neural networks in generating radar echo frames also gains much attention recently. In general, convolutional neural network (CNN) is employed to capture spatial dependencies while recurrent neural network (RNN) is aimed at capturing temporal dependencies. And existing works can be divided into two classes: pure convolutional neural networks  \cite{agrawal2019machine,TREBING2021178} and convolutional recurrent neural networks \cite{xingjian2015convolutional,wang2022predrnn,shi2017deep,zhong2020spatiotemporal}.
 However, CNN suffers from the inductive bias (i.e., translation invariance and locality), which cannot capture location-variant information (i.e., natural motion and transformation) and is weak at extracting long-range dependencies. As for RNN, it is time-consuming in the process of long back-propagation  due to the seriality of recurrent structure.

Aside from CNN and RNN, Transformer-based methods also gain more and more attention due to the great potential of Transformer structure \cite{bai2022rainformer}. With powerful multi-head attention mechanism, Transformer-based methods can capture long-term dependencies efficiently and be implemented in parallel, which achieves good performance and fast training speed. However, these works rarely consider short-term dependencies and are implemented in a non-autoregressive way, which is hard to generate consecutive and accurate radar echo frames.

In this paper, we propose a variant of Transformer named patches with 3D-temporal convolutional Transformer network (PTCT), where original frames are split into multiple patches for removing the constraint of inductive bias and 3D-temporal convolutional operation is employed to capture short-term dependencies efficiently.
Briefly speaking, PTCT consists of three components: patch embedding, spatiotemporal encoder and frame forecaster. Patch embedding aims at removing the constraint of inductive bias to provide better feature representation for spatiotemporal encoder. Then, spatiotemporal encoder captures spatial and temporal dependencies, especially short-term dependencies. Finally, frame forecaster generates the predicted radar echo frames. To ensure the quality of radar echo frames generated, we adopt an autoregressive way during inference of PTCT, where the newest radar echo frame is appended to previous ones to generate the next radar echo frame.

To validate our proposed algorithm, we conduct experiments on two radar echo dataset: Radar Echo Guangzhou and Hong Kong Observatory (HKO-7). The experimental results show that PTCT achieves state-of-the-art (SOTA) performance compared with existing methods. Numerous ablation experiments are also conducted to reveal effectiveness of different components

\section{Related Work}
In this section, we list related works with neural networks in precipitation nowcasting since neural networks usually perform better than traditional methods. We mainly divided these works into three classes: pure convolutional neural networks,  convolutional recurrent neural networks and Transformer-based networks.

We begin by introducing pure convolutional neural networks. The ubiquitous U-Net convolutional neural network transfers well-known UNet architecture into precipitation nowcasting, where convolutional operations are used to capture spatial dependencies and  multiple radar frames are stacked to form the temporal dimension for extracting temporal dependencies \cite{agrawal2019machine}. To reduce the model parameter size of the original UNet, Small Attention-UNet (SmaAt-UNet) equips UNet with attention modules and depthwise-separable convolutions \cite{TREBING2021178}. However, the inherent inductive bias constrains their ability for capturing location-invariant information and long-range dependencies, and the lack of autoregressive characteristic makes them hard to track long-time radar echo changes effectively.

As for recurrent convolutional neural networks, convolutional LSTM (ConvLSTM) establishes a seminal framework for precipitation nowcasting, where convolutional neural network and recurrent neural network are combined to capture both spatial dependencies and temporal dependencies, respectively \cite{xingjian2015convolutional}. To remove the limit of location-invariant convolution filters in ConvLSTM, the Trajectory Gated Recurrent Unit (TrajGRU) uses a subnetwork to output the state-to-state connection structures before state transitions, which can actively learn the location-variant structure for
recurrent connections \cite{shi2017deep}. Predictive recurrent neural network (PredRNN) designs a zigzag memory flow that propagates in both bottom-up and top-down direction across all layers, which aims to communicate the learned visual dynamics at different levels \cite{wang2017predrnn}. The improved predictive recurrent neural network (PredRNN++) proposes a new recurrent structure named Causal LSTM for modeling short-term dependencies, which adds more non-linear layers to recurrent transition and employs a gradient highway unit (GHU) to alleviate the vanishing gradient problem \cite{wang2018predrnn++}. EIDETIC 3D LSTM (E3D-LSTM) introduces 3D-convolution into the LSTM cell to capture short-term dependencies more efficiently \cite{wang2018eidetic}. PredRNN-V2 further improves PredRNN by decoupling the interlayer spatiotemporal memory and inner-layer temporal memory in laten space and proposes a new curriculum learning strategy \cite{wang2022predrnn}.
Spatiotemporal convolutional long short-term memory (ST-ConvLSTM) adds an attention block into the LSTM cell so as to model long-range and long-term spatiotemporal dependencies \cite{zhong2020spatiotemporal}. Similarly, self-attention ConvLSTM (SA-ConvLSTM) employs a more compilcated self-attention memory into the LSTM cell to capture long-range dependencies efficiently \cite{lin2020self}. Though numerous variants of ConvLSTM have been developed, these methods suffer from heavy computation burden such as time and long back-propagation process due to the seriality of recurrent structure.

Transformer, originally proposed in natural language processing (NLP), has been transferred to many other areas due to its great potential for extracting rich long-term dependencies and good parallelism \cite{vaswani2017attention}. Based on the window-based multi-head self-attention mechanism of SwinTransformer \cite{liu2021swin}, Rainformer extracts robust global features and employs the gate fusion unit to balance local and global features for high-intensity rainfall prediction \cite{bai2022rainformer}. Another relevant work is Convolutional Transformer (ConvTransformer), which stacks multiple Transformer blocks in the encoder and decoder directly to capture the temporal dependencies \cite{liu2020convtransformer}. However, both Transformer-based methods rarely consider short-term dependencies and generate future frames in a non-autoregressive way. In addition, the inherent inductive bias of convolution still exists in these Transformer-based methods.

Aside from related works in precipitation nowcasting, some researches about patches in computer vision also inspire us a lot. Vision Transformer (ViT) splits an image to fixed-size patches and linearly embed each of them as the input of Transformer block, which performs well on image classification tasks \cite{dosovitskiy2020image}. Masked autoencoder (MAE)  masks random patches of the input image and then reconstruct the mixing pixels using an encoder-decoder architecture, which reveals the great potential of masking in feature extraction\cite{he2021masked}. ConvMixer removes the self-attention layer of ViT and provides an evidence that using patch as the input representation plays a more important role in the good performances of ViT\cite{dosovitskiy2020image}. In summary, these works strengthen the importance of patches for image processing. Actually, ConvLSTM, PredRNN and ST-ConvLSTM also split original frames into a few patches, but they regard it as a way to save computation resources. We go even further in the research of patches for precipitation nowcasting.

Thus, observations above motivate us to develop an efficient variant of Transformer for precipitation nowcasting, which aims at removing the constraint of inductive bias of convolution and extracting short-term dependencies efficiently. In addition, the autoregressive characteristic is kept during inference to ensure the quality of future frames generated.

\section{Patches with 3D-Temporal Convolutional Transformer Network (PTCT)}
\begin{figure*}[htbp]
	\centering
\label{fig:overview}
\includegraphics[width=1.0\linewidth, height=4.5in]{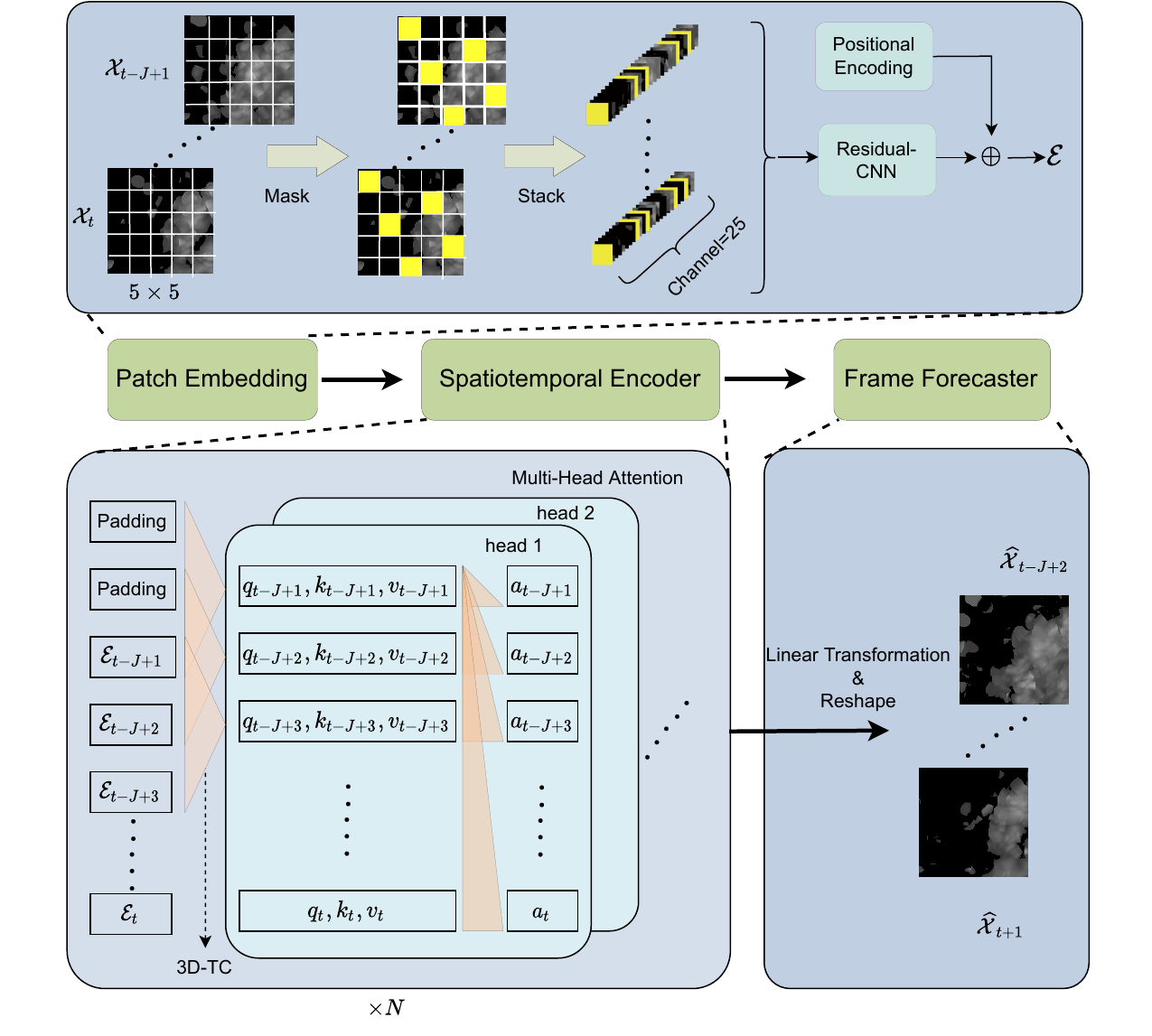}
	\caption{The overview of PTCT is shown at time $t$. In the patch embedding, $\mathcal{X}_t\in\mathbb{R}^{140\times140\times1}$ are first masked and then split into $5\times 5$ patches for stack. In the spatiotemporal encoder, 3D-TC is first performed and then multi-head attention is employed. Along the temporal dimension,  $q_{i},k_{i},v_{i}$ and $a_i$ represent the $i$-th element of $Q,K,V$ and $\mathcal{A}$, respectively. Finally, the frame forecaster uses a linear transformation to generate the predicted frame.}
\end{figure*}

In this section, we first demonstrate the problem setup of precipitation nowcasting. Then, we introduce three components of PTCT, i.e., patch embedding, spatiotemporal encoder and frame forecaster. Patch embedding aims at removing the constraint of inductive bias to provide better feature representation for spatiotemporal encoder. Then, spatiotemporal encoder captures spatial and temporal dependencies, especially short-term dependencies. Finally, frame forecaster generates the predicted radar echo frames. The overview of PTCT is shown in Figure \ref{fig:overview}.

\subsection{Problem Setup}
Precipitation nowcasting is to generate future frames of radar echo given a sequence of historical radar echo frames. Formally, given a frame sequence of length $J$ denoted as $\mathcal{X}=\{\mathcal{X}_{t-J+1},\dots,\mathcal{X}_{t}\}$ where $\mathcal{X}_t\in\mathbb{R}^{H\times W\times C}$ is the radar echo frame with $H$-height, $W$-width and $C$-channel at time t, the goal is to generate the most probable length-$L$ sequence in the future:
\begin{equation}\label{eq:problem-definition}
  \widehat{\mathcal{X}}_{t+1},\dots,\widehat{\mathcal{X}}_{t+L}=\mathop{\arg\max}_{\mathcal{X}_{t+1},\dots,\mathcal{X}_{t+L}} \mathbb{P}(\mathcal{X}_{t+1},\dots,\mathcal{X}_{t+L}|\mathcal{X}),
\end{equation}
where $\mathbb{P}(\cdot|\cdot)$ represents the conditional probability.  We define the predicted sequence as $\widehat{\mathcal{X}}=\{\widehat{\mathcal{X}}_{t+1},\dots,\widehat{\mathcal{X}}_{t+L}\}$ for notational simplicity.


\subsection{Patch Embedding}

 To remove the limit of inductive bias of convolution in precipitation nowcasting, we design an elaborate patch embedding block to extract long-range dependencies and preserve location-variant spatial features. The core idea is to divide the original radar echo frame into multiple patches and stack them as the input of convolutional neural network.

Given historical frames $\mathcal{X}\in\mathbb{R}^{J\times H\times W\times C}$, we first split them into $m\times n$ patches for each frame, denoted as $\mathcal{X}^p\in\mathbb{R}^{J\times\frac{H}{m}\times\frac{W}{n}\times( m\cdot n\cdot C)}$. Then, we can perform convolution directly to capture both short-range and long-range dependencies, with location-variant spatial features extracted. Concretely, we employ a residual-CNN as given in Eq.(\ref{eq:cnn-embedding}):

\begin{equation}\label{eq:cnn-embedding}
  \begin{aligned}
  \mathcal{G}&=\sigma(W_{\mathcal{G}}*\mathcal{X}^p),\\
  \mathcal{H}&=\sigma(W_{\mathcal{H}}*\mathcal{G})+\mathcal{G},
  \end{aligned}
\end{equation}
where $\mathcal{H}\in\mathbb{R}^{J\times \frac{H}{m}\times \frac{W}{n}\times D}$ represents feature maps with $D$ output channels, $W_{\mathcal{G}}$ and $W_{\mathcal{H}}$ are 2D-convolutional kernels, $\sigma(\cdot)$ represents the leaky rectified linear unit (LReLU) activation function and $*$ denotes 2D-convolution operation. Note that the way of our patch representation is different from the one in ViT \cite{dosovitskiy2020image}, which linearly embeds patches as the input representation and thus results in a heavy burden of computational resource for high-resolution radar echo frames. Instead, we employ the convolution to extract features of patches, which reducing the number of parameters greatly.

To explain how patch embedding takes effect intuitively, we take one convolutional kernel of $W_{\mathcal{G}}$ and $C=1$ as an example. In practice, a convolutional kernel consists of input channel $\mathsf{C}_{in}$, height $\mathsf{h}$ and width $\mathsf{w}$. Here, it is easy to know that $\mathsf{C}_{in}=C\cdot m\cdot n=m\cdot n$. Further, we subdivide the convolutional kernel as $m\times n$ learnable weight matrix $W_i\in\mathbb{R}^{\mathsf{h}\times\mathsf{w}},\forall i\in\{1,\dots,m\cdot n\}$. Then, it is ready to explain as follows:
\begin{itemize}
  \item  Since patches are stacked together, spatial features at different patches can be aggregated simultaneously through the convolution operation, which enriches spatial dependencies with long-range ones.
  \item For patch $i$, $W_i$ is the same across all positions for capturing location-invariant spatial features. However, $W_j$ of patch $j$ is different from $W_i$ to ensure no mixture of location-variant spatial features. Thus, there is no interference between any two patches, which makes both location-invariant and location-variant spatial features satisfied at the same time.
\end{itemize}
In this way, long-range dependencies and location-variant spatial features are added successfully, which helps to remove the limit of inductive bias of convolution.

Meanwhile, motivated by the great potential of masking in \cite{he2021masked}, we mask random patches of original frames and reconstruct them in the loss function. It is also beneficial for the case that there are many empty values in real-world radar echo frames. We also regard it as a kind of "dropout" mechanism \cite{srivastava2014dropout}, which is helpful to avoid overfitting.

Finally, we employ a regular positional encoding to form the input $\mathcal{E}\in\mathbb{R}^{J\times \frac{H}{m}\times \frac{W}{n}\times D}$of the following spatiotemporal encoder, which is given as follow:
\begin{equation}\label{eq:embedding-final}
  \mathcal{E} = \mathcal{H}\oplus \mathcal{P},
\end{equation}
where $\oplus$ represents the element-wise addition and $\mathcal{P}$ is the positional encoding as defined in \cite{vaswani2017attention}.

\subsection{Spatiotemporal Encoder}
In precipiation nowcasting, there exist strong correlation between adjacent radar echo frames, indicating that short-term dependencies plays a crucial role. Thus, it is necessary to strengthen short-term dependencies during learning. To be specific, we introduce the 3D-temporal convolution (3D-TC) into the self-attention layer of Transformer block, instead of  plain linear transformation given in \cite{vaswani2017attention}. At each Transformer block, we divide it into two components: multi-head attention and feed forward.

\textbf{Multi-head Attention.} At first, we employ a Pre-Layer Normalization such that $\widehat{\mathcal{E}}=\text{Pre-LN}(\mathcal{E})$ as given in \cite{xiong2020layer}. Then, the core equations of TC Self-Attention are given as follows:
\begin{align}
	Q = W_{Q}*\widehat{\mathcal{E}},\qquad
	K = W_{K}*\widehat{\mathcal{E}},\qquad
	V = W_{V}*\widehat{\mathcal{E}},
\end{align}
where $Q,K,V\in\mathbb{R}^{J\times H\times W\times D}$ are the query, key and value, respectively. $W_Q$, $W_K$ and $W_V$ represent the trainable 3D-temporal convolutional kernels. With slight abuse of notation, we also denote $*$ as the 3D-temporal convolution operation. As shown in Figure \ref{fig:overview}, 3D-temporal convolution takes adjacent frames as input to strengthen the short-term dependencies more explicitly. Meanwhile, the intrinsic property that  considering only previous frames makes 3D-temporal convolution free of future frames leaking and naturally suit for the following masked self-attention:
\begin{equation}\label{eq:masked-self-attention}
    \begin{aligned}
  \mathcal{A}=\text{Softmax}\bigg(\text{SeqMask}(\frac{QK^T}{\sqrt{D}})\bigg)V,
    \end{aligned}
\end{equation}
where $\text{SeqMask}(\cdot)$ is the sequence mask to prevent future information from leaking. This way, long-term dependencies can be extracted for precipitaion nowcasting. Further, with multiple outputs of TC Self-Attention concatenated, a linear transformation is performed to change the representation space. This way, it is beneficial to extract abundant features from different representation space, known as multi-head attention. We denote the output of multi-head attention as $\mathcal{M}$. Thereafter, a regular short-cut is employed such that $\mathcal{N}=\mathcal{E}\oplus\mathcal{M}$.

\textbf{Feed Forward.} At first, we perform a regular Pre-Layer Normalization as $\widehat{\mathcal{N}}=\text{Pre-LN}(\mathcal{N})$  and then send it to a feed forward network such that $\mathcal{F}=\text{FFN}(\widehat{\mathcal{N}})$. After that, a regular short-cut is performed as the final output of a Transformer block, i.e., $\mathcal{O}=\mathcal{N}\oplus\mathcal{F}$.

Finally, we stack $N$ Transformer blocks to enhance the representation ability of our model. We denote the output of the final Transformer block as $\mathcal{Z}\in\mathbb{R}^{J\times \frac{H}{m}\times \frac{W}{n}\times D}$. Note that during the training process, we also apply dropout technique to the output of each sub-layer, including the attention score produced by $\text{Softmax}(\cdot)$.

\subsection{Frame Forecaster}
With historical length-$J$ frames, we aim to generate the most probable length-$L$ frames. It can be achieved by a linear transformation and a reshape patch back operation. Linear transformation is given in Eq.(\ref{eq:forcast}):
\begin{equation}\label{eq:forcast}
  \widehat{\mathcal{X}}^p=\mathcal{Z}W_{\widehat{\mathcal{X}}^p},
\end{equation}
where $W_{\widehat{\mathcal{X}}^p}$ represents the learnable weight. And then $\widehat{\mathcal{X}}^p$ is reshaped back to $\widehat{\mathcal{X}}\in\mathbb{R}^{J\times H\times W\times C}$  . The model is optimized by minimizing the loss function is as follow:
\begin{equation}\label{eq:loss=function}
  Loss = \frac{1}{J+L-1}\sum_{i=t-J+2}^{t+L}l (\mathcal{X}_i,\widehat{\mathcal{X}}_i)
\end{equation}
where $l(\cdot,\cdot)$ can be any pre-defined function measuring the similarity of two images, e.g., mean square error (MSE).

To ensure the quality of future frames generated, we do not generate  $\{\mathcal{X}_{t+1},\mathcal{X}_{t+L}\}$ directly according to $\{\mathcal{X}_{t-J+1},\mathcal{X}_{t}\}$ during the process of inference, which is adopted in \cite{liu2020convtransformer}. Instead, we generate future frames in an autoregressive way. To be specific, during the training process, we still take $\{\mathcal{X}_{t-J+1},\dots,\mathcal{X}_{t+L-1}\}$ as input to generate the most probable $\{\widehat{\mathcal{X}}_{t-J+2}\dots,\widehat{\mathcal{X}}_{t+L}\}$ for calculating loss function. However, during the process of inference, we generate $\widehat{\mathcal{X}}$ frame by frame autoregressively. Concretely, we take $\{\mathcal{X}_{t-J+1},\dots,\mathcal{X}_{t}\}$ to generate $\widehat{\mathcal{X}}_{t+1}$ and concatenate it to form $\{\mathcal{X}_{t-J+1},\dots,\mathcal{X}_{t},\widehat{\mathcal{X}}_{t+1}\}$ as the next input. After applying it repeatedly, we can generate the future length-$L$ frames successfully.

\section{Experiment}
To validate our proposed algorithm, we conduct experiments over two dataset: Radar Echo Guangzhou  \cite{zhong2020spatiotemporal} and HKO-7 \cite{shi2017deep}.

\textbf{Dataset.} Radar Echo Guangzhou dataset consists of 4088 consecutive radar observations recorded every 12 minutes at Guangzhou, China.  Further, these radar observations are split into 3280, 408 and 400 sequences for training, validating and testing, respectively. Each sequence contains 10 input frames and 10 output frames standing for the past 2 hours and future 2 hours, respectively. The size of each radar echo map is $140\times140$ with spatial resolution as 1 kilometer.

HKO-7 includes 20949 consecutive radar observations recorded every 6 minutes at HongKong, China. There are 18177, 808 and 1964 sequences for training, validating and testing, respectively. Similarly, each sequence contains 10 input frames and 10 output frames representing the past 1 hours and future 1 hours, respectively. The size of each radar echo map is $480\times480$ with spatial resolution as 1.07 kilometer, which is more complicated and challenging.

\textbf{Evaluation Metrics.} We adopt three evaluation metrics commonly used in precipitation nowcasting, i.e, structural similarity index measure (SSIM), critical success index (CSI) and probability of detection (POD). SSIM measures how similar the predicted frames are with ground truth in the structural aspect. Since the generated frames are used to predict if rainy or not, we  set a threshold as 40dBZ. If the value of a pixel is greater than or equal to 40dBZ, we label it as true, otherwise false. Further, we denote $TP$ as true positive, $FP$ as false positive and $FN$ as false negative to measure difference between the generated frames and the true frames. CSI is calculated as $\frac{TP}{TP+FN+FP}$. The higher CSI is, the better the prediction of the radar echo reflectivity is. POD is calculated as $\frac{TP}{TP+FN}$, which represents the accuracy of issuing a warning for exceeding the threshold.

\textbf{Baselines.} Baselines include the traditional method in computer vision (i.e., dense inverse search (DIS) optical flow \cite{kroeger2016fast}), convolutional recurrent neural networks (i.e., ConvLSTM \cite{xingjian2015convolutional}, SA-ConvLSTM \cite{lin2020self}, ST-ConvLSTM \cite{zhong2020spatiotemporal}, PredRNN-V2 \cite{wang2022predrnn} and TrajGRU \cite{shi2017deep}) and Transformer-based network (i.e., ConvTransformer \cite{liu2020convtransformer}). PredRNN-V2 are regarded as state-of-the-art in precipitation nowcasting currently.

\textbf{Implementation Details.}
We conduct all the experiments with Intel Xeon(R) Gold 6132 CPU @2.60GHz, four NVIDIA V100 GPUs and 240GB memory. The operating system is CentOS 7.4.1708. For baseline methods, we implement them with hyperparamters given in their papers. For PTCT, six layers or blocks with $D=800$. Further, the size of 3D-temporal convolutional kernel is $3\times3\times3$. We set the mini-batch as 8 for all the methods in Radar Echo GuangZhou, and 4 for HKO-7 dataset.

To optimize models, we select mean square error (MSE) as the loss function for all the methods. We further adopt the ADAM optimizer with  learning rate as $10^{-5}$. Further, the cosine annealing schedule\cite{loshchilov2016sgdr} is employed to train PTCT while baseline methods are trained as described in their papers. And we train PTCT for 630 and 41 epoches in Radar Echo GuangZhou and HKO-7, respectively.

\subsection{Results}

\begin{wraptable}{r}{2.8in}
    \vspace{-2em}
	\centering
	\caption{Performances on the Radar Echo Guangzhou.}
    \label{table:radar-gz}
	\begin{tabular}{lccccc}
		\toprule
		\multicolumn{1}{c}{MODEL}&
		\multicolumn{1}{c}{SSIM }&
		\multicolumn{1}{c}{CSI }&
		\multicolumn{1}{c}{POD }\\
		\midrule
		DIS optical flow \cite{kroeger2016fast} & 0.434 & 0.304 & 0.451\\
		ConvLSTM \cite{xingjian2015convolutional} & 0.632 & 0.232 & 0.269\\
		TrajGRU \cite{shi2017deep} & 0.628 & 0.244 & 0.289\\
		PredRNN \cite{wang2022predrnn} & 0.630 & 0.273 & 0.331\\
		ST-ConvLSTM \cite{zhong2020spatiotemporal} & 0.655 & 0.282 & 0.335\\
		SA-ConvLSTM \cite{lin2020self} & 0.658 & 0.280 & 0.336\\
		ConvTransformer \cite{liu2020convtransformer} & 0.163 & 0.009 & 0.015\\
		PredRNN-v2 \cite{wang2022predrnn} & 0.692 & 0.316 & 0.382\\
		\midrule
        PTCT  & \textbf{0.907} & \textbf{0.531} & \textbf{0.701}\\

		\hline
	\end{tabular}
\vspace{-1em}
\end{wraptable}

\begin{figure*}[htbp]
\centering
	
	\includegraphics[width=0.32\linewidth]{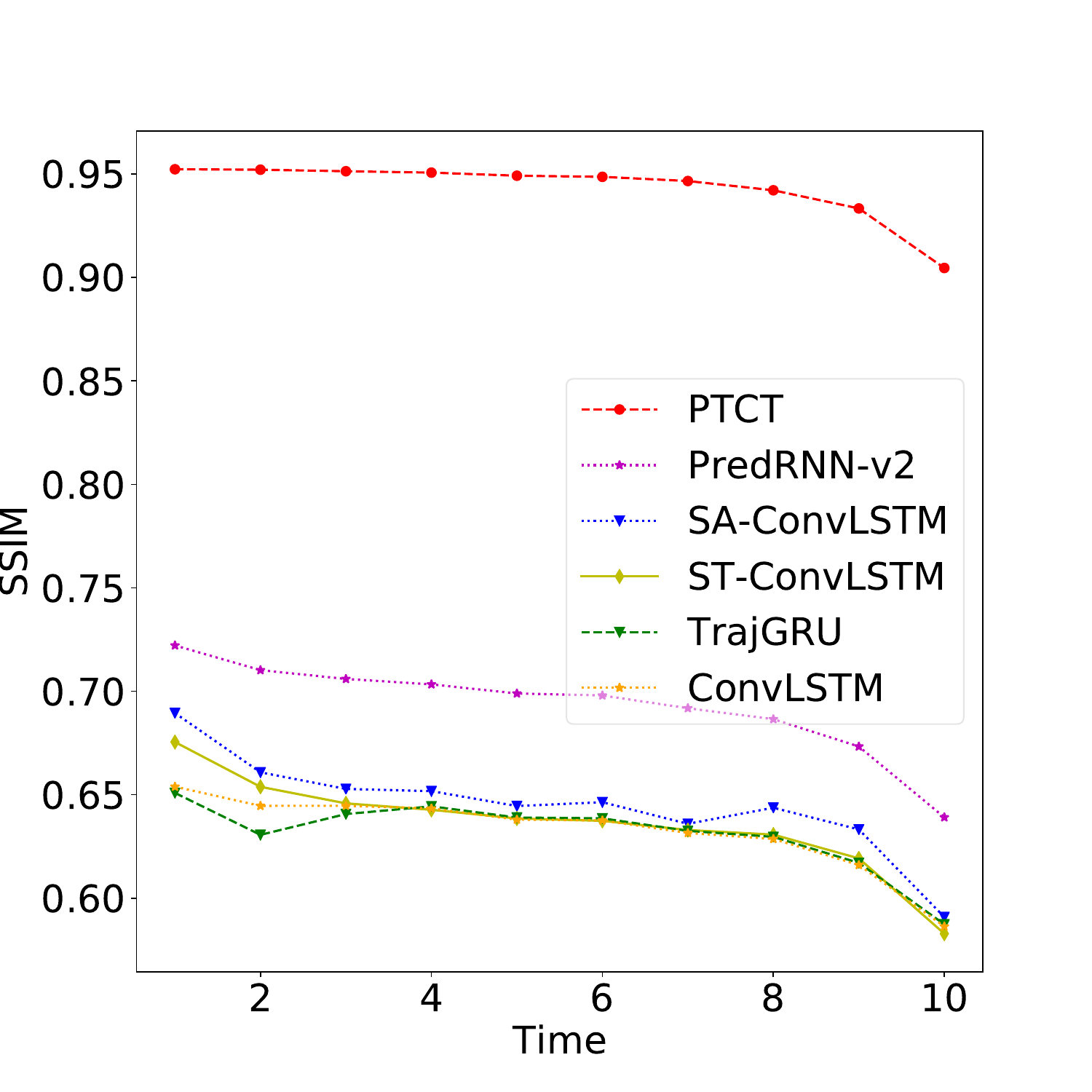}
	\includegraphics[width=0.32\linewidth]{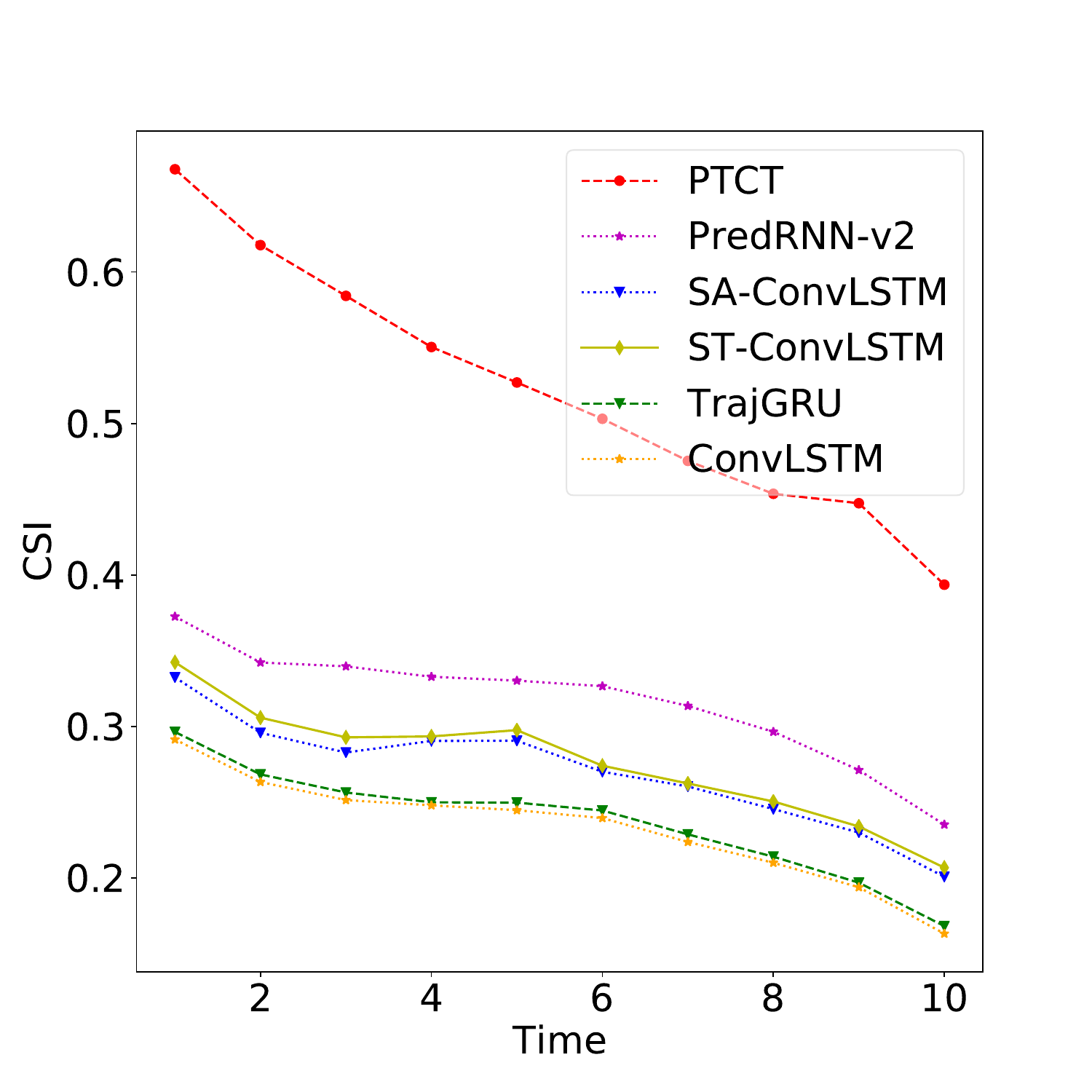}
	\includegraphics[width=0.32\linewidth]{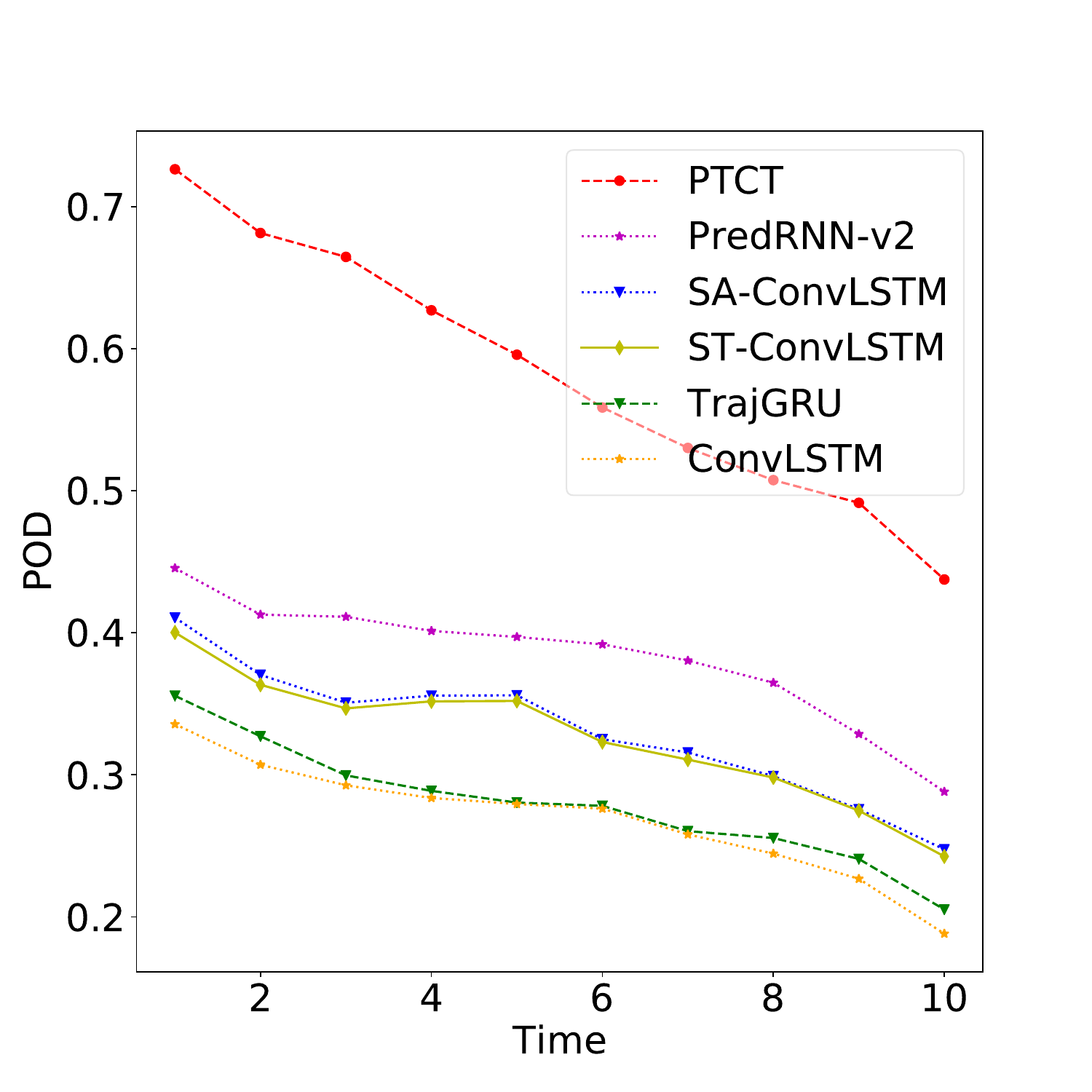}
\caption{Frame-wise SSIM, CSI and POD on Radar Echo Guangzhou.}\label{fig:frame-wise-radar}
\end{figure*}

\textbf{Results on Radar Echo Guangzhou.} We report the results of Radar Echo Guangzhou
in Table \ref{table:radar-gz}. Compared with baselines, our proposed PTCT outperforms them significantly in all the metrics, especially in SSIM and POD. We further plot frame-wise metrics in Figure \ref{fig:frame-wise-radar},  where the x-axis represents the time step of future 10 frames and the y-axis is the metric for measuring the performance of predicted frames.

\begin{figure}[!htbp]
	\centering
	\includegraphics[width=0.8\linewidth]{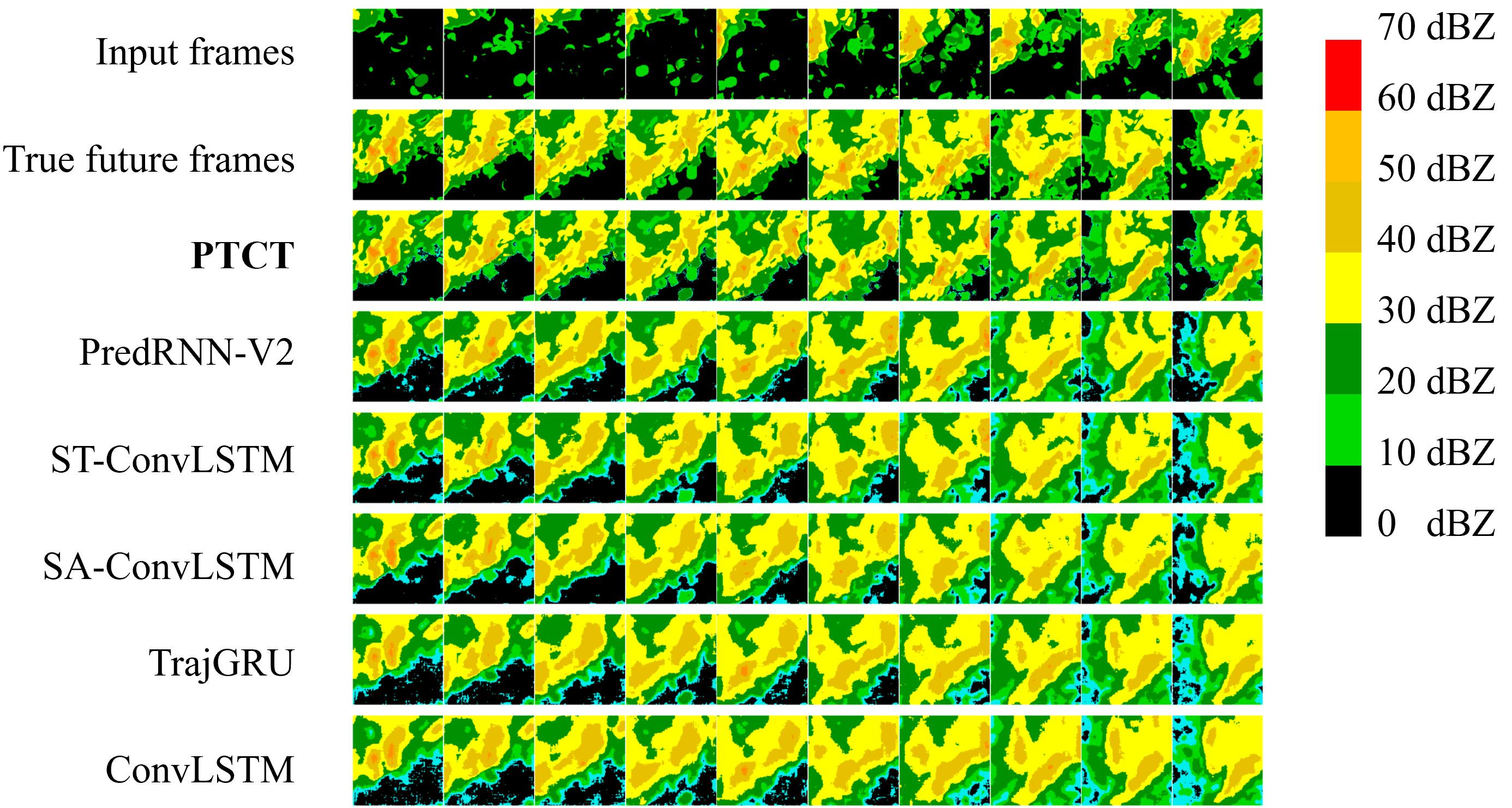}
    \caption{Predicted frames on Radar Echo Guangzhou.}\label{fig:prediction-radar-seq}
\end{figure}

We further visualize the predicted radar frames by mapping them into RGB space, where areas with values larger than 40 dBZ tend to be more rainy. As shown in Figure \ref{fig:prediction-radar-seq}, PTCT generates the best predicted frames in all metrics all the time, which validates the effectiveness of PTCT.

\begin{wraptable}{r}{2.4in}
    \centering
    \vspace{-2em}
	\caption{Performances on the HKO-7.}
    \label{table:radar-hko}
	\begin{tabular}{lccccc}
		\toprule
		\multicolumn{1}{c}{MODEL}&
		\multicolumn{1}{c}{SSIM }&
		\multicolumn{1}{c}{CSI }&
		\multicolumn{1}{c}{POD }&\\
		\midrule
		ConvLSTM \cite{xingjian2015convolutional} & 0.786 & 0.199 & 0.224 \\
		TrajGRU \cite{shi2017deep} & 0.769 & 0.203 & 0.231 \\
		ST-ConvLSTM \cite{zhong2020spatiotemporal} & 0.762 & 0.147 & 0.165 \\
		SA-ConvLSTM \cite{lin2020self} & 0.761 & 0.148 & 0.169\\
		PredRNN-v2 \cite{wang2022predrnn} & 0.770 & 0.162 & 0.181 \\
		\midrule
        PTCT  & \textbf{0.796} & \textbf{0.243} & \textbf{0.409}\\
		\hline
	\end{tabular}%
    \vspace{-1em}
\end{wraptable}

To exhibit performances of difference methods directly, we sample a sequence from the test set and plot the predicted frames of different methods in Figure \ref{fig:prediction-radar-seq}. Obviously, PTCT keeps the clearest and  the most similar structure of radar echo as the true future frames especially in the last two frames, where frames generated by other methods  are too blurry to track the dynamics exactly.

\textbf{Results on HKO-7.} For more complicated and challenging HKO-7 dataset, we present the results in Table \ref{table:radar-hko}. Compared with baselines, PTCT still outperforms them in SSIM, CSI and POD.

In summary, PTCT achieves state-of-the-art (SOTA) performances on Radar Echo Guangzhou and HKO-7.

\subsection{Ablation study}
To reveal effectiveness of patch embedding and 3D-temporal convolution, numerous ablation experiments are conducted on Radar Echo Guangzhou.

\begin{wraptable}{r}{2.6in}
    \centering
    \vspace{-2em}
	\caption{Performances of different patch sizes.}
    \label{table:radar-patch}%
	\begin{tabular}{lcccc}
		\toprule
		\multicolumn{1}{c}{Patch Size} &
		\multicolumn{1}{c}{SSIM} &
		\multicolumn{1}{c}{CSI}&
		\multicolumn{1}{c}{POD} \\
		\midrule
		$1\times1$ &
		 0.203 & 0.015 & 0.019 \\
		
		$5\times5$   & 0.419 & 0.078 & 0.092 \\
		
		$14\times14$   & 0.545 & 0.238 & 0.278 \\
		\midrule
		$20\times20$  & \textbf{0.907} & \textbf{0.531} & \textbf{0.701} \\

		\midrule
		
	\end{tabular}%
    \vspace{-2em}
\end{wraptable}

\textbf{Patch \& Mask.} To investigate the influence of different patch sizes, we test it with $1\times1$, $5\times5$, $10\times10$, $14\times 14$ and $20\times20$. The case of $1\times1$ means no split of original frames. All the results are presented in Table \ref{table:radar-patch}. As patch size grows, the performance is improved gradually, which means splitting original frames into multiple patches is essential to remove the constraint of inductive bias.

\begin{table*}[!h]
	\centering
	\caption{Performances of different cases of location-variant features and long-range dependencies.}
    \label{table:translation}%
	\begin{tabular}{ccccccccc}
		\toprule
		\multicolumn{1}{c}{location-variant} &
		\multicolumn{1}{c}{long-range} &
		\multicolumn{1}{c}{Patch Size} &
		\multicolumn{1}{c}{Groups} &
        \multicolumn{1}{c}{Heads} &
		\multicolumn{1}{c}{SSIM} &
		\multicolumn{1}{c}{CSI}&
		\multicolumn{1}{c}{POD} \\
		\midrule
		
		\XSolid & \XSolid & $1\times1$ & 1 & 8 & 0.203 & 0.015 & 0.019 \\
		\checkmark & \XSolid & $20\times20$ & 400 & 400 & 0.331 & 0.022 & 0.026 \\
		\checkmark & \checkmark & $20\times20$ & 1 & 8 & \textbf{0.907} & \textbf{0.531} & \textbf{0.701} \\

		\midrule
		
	\end{tabular}%
\end{table*}

To further reveal the importance of location-variant spatial features and long-range dependencies, we test it by setting different group sizes of group convolution, and the head number of multi-head attention. All the results are given in Table \ref{table:translation}. The second row indicates that location-variant information is beneficial for improving the performance. The third row reveals that none of the location-variant spatial features and long-range dependencies can be omitted.

\begin{wraptable}{r}{3.0in}
    \centering
    \vspace{-1.8em}
    \caption{Performances of different ratios of mask under training and testing phrases. }
    \label{table:radar-mask}%
	\begin{tabular}{lccccc}
		\toprule
		\multicolumn{1}{c}{Training} &
		\multicolumn{1}{c}{Testing} &
		\multicolumn{1}{c}{SSIM} &
		\multicolumn{1}{c}{CSI}&
		\multicolumn{1}{c}{POD} \\
		\midrule
		0\% & 0\% & \textbf{0.946} & 0.522 & 0.582\\
		
		25\% & 0\% & 0.907 & \textbf{0.531} & 0.701\\
		
		50\% & 0\% & 0.885 & 0.505 & \textbf{0.726}\\
		\midrule
		
		25\% & 25\% & 0.829 & 0.414 & 0.582\\
		
		50\% & 50\% & 0.813 & 0.389 & 0.553\\

		\midrule
		
	\end{tabular}%

\end{wraptable}
The mechanism of masking is an important component of patch embedding. To investigate the influence of masking patches, we test it using different ratios of mask. We also apply mask mechanism during inference to show great representation ability of our model.
All the results are given in Table \ref{table:radar-mask}. As the ratio of mask grows, it degrades performances in SSIM while improving performances in POD. Meanwhile, appropriate ratio of mask is important for CSI. Thus, there exist a tradeoff among different metrics. Finally, the last two rows of Table \ref{table:radar-mask} reveal that relatively good performance can be guaranteed even if half of the input frames are masked. We believe that PTCT has learned how to reconstruct the masked-out patches and thus remain relatively good performance, which can be beneficial for the case of noisy radar echo frames (e.g., due to the limit of meteorological devices).

\begin{wraptable}{r}{3.0in}
    \centering
    \vspace{-2em}
    \caption{Performances of 2D convolution and 3D-temporal convolution. }
    \label{table:radar-3D}%
	\begin{tabular}{lcccc}
		\toprule
		\multicolumn{1}{c}{Kernel Size} &
		\multicolumn{1}{c}{SSIM} &
		\multicolumn{1}{c}{CSI40}&
		\multicolumn{1}{c}{POD40} \\
		\midrule
		
		$3\times3\times3$ (3D) & \textbf{0.907} & \textbf{0.531} & \textbf{0.701}  \\
		
		$3\times3$ (2D)  & 0.843 & 0.467 & 0.551 \\

		\midrule
		
	\end{tabular}%
    \vspace{-1em}
\end{wraptable}
\textbf{3D-Temporal Convolution.} We replace 3D-temporal convolution with 2D convolution, which reduces the ability of capturing short-term dependencies. Results are presented in Table \ref{table:radar-3D}, where 3D-temporal convolution outperforms 2D-convolution in all metrics.

Due to the limit of page length, more details and results of our experiments are available in the supplementary. The code will be released at github after being accepted.


\section{Conclusion}

In this paper, we propose the pacthes with 3D-temporal convolutional Transformer network (PTCT) for precipitation nowcasting. PTCT first splits original frames into multiple patches to remove the constraint of inductive bias, and then incorporate 3D-temporal convolution within the self-attention mechanism to capture short-term dependencies more efficiently. The experimental results show effectiveness of PTCT, which achieves SOTA performances on two radar echo dataset. Numerous ablation experiments  are  conducted to reveal effectiveness of different components.

In future, we plan to extend our work by training a pre-trained model as a foundation model for cross-region tasks, which saves the training time and achieves a better performance through fine-tuning.

\begin{ack}
This work is supported in part by National Natural Science Foundation of China under grant U1811464, Science and Technology Program of Guangzhou under grant 201903010104, and Guangdong Provincial Key Laboratory of Computational Science at Sun Yat-Sen University under grant 2020B1212060032.

We express our great gratitude to Prof. Qing Ling and Shuxin Zhong for their patient and detailed discussions of our work.

\end{ack}
%
%
%
%
%
%
%
%
%
%



\appendix

\bibliographystyle{plainnat}
\bibliography{ref_backup}

\clearpage

\section*{Appendex}

\setcounter{table}{6}
\setcounter{figure}{3}

\appendix
\renewcommand{\appendixname}{Appendix~\Alph{section}}

\section{HyperParameters}
In this section, we mainly list some critical hyperparameters of different models, as shown in Table \ref{table:parame-guangzhou} and \ref{table:parame-hko-7}. Note that previous models still split their original frames into patches for saving computational resource. Here, we set the patch size of them as $4\times4$ adopted in their papers. Meanwhile, previous models usually stack multiple sub-layers to enhance their representation ability, which is also a critical hyperparameter. For simplicity, we also call the number of Transformer blocks in PTCT as the number of layers. Other hyperparameters are set as adopted in the original papers of these models. We also calculate the amount of parameters of different models.

As shown in Table \ref{table:parame-guangzhou}, as the patch size grows, the parameter amount of PTCT also increases significantly, which relies heavily the ratio between patch size and $D$.

\begin{table}[!h]
\centering
\caption{HyperParameters of Different Models on Radar Echo Guangzhou.}
\label{table:parame-guangzhou}%
\begin{tabular}{lccccc}
	\toprule
	\multicolumn{1}{c}{MODEL}&
	\multicolumn{1}{c}{Patch Size }&
	\multicolumn{1}{c}{$D$}&
	\multicolumn{1}{c}{Layers }&
	\multicolumn{1}{c}{Parameter Amount }&\\
	\midrule
	ConvLSTM & $4\times4$ & 128 &4& 18.87M \\
	TrajGRU & $4\times4$ & 128 &4& 20.78M\\
	ST-ConvLSTM & $4\times4$ & 128 &4&  42.28M \\
	SA-ConvLSTM & $4\times4$ & 128 &4& 33.70M \\
	PredRNN-v2 & $4\times4$ & 128 &4& 42.93M \\
	\midrule
	PTCT & $1\times1$ & 2 &6& 0.31M \\
	PTCT & $5\times5$ & 50 &6& 1.87M \\
	PTCT & $14\times14$ & 392 &6& 144.65M \\
    PTCT & $20\times20$ & 800 &6& 488.48M \\
	\hline
\end{tabular}%
\end{table}%

\begin{table}[!h]
\centering
\caption{HyperParameters of Different Models on HKO-7.}
\label{table:parame-hko-7}%
\begin{tabular}{lccccc}
	\toprule
	\multicolumn{1}{c}{MODEL}&
	\multicolumn{1}{c}{Patch Size }&
	\multicolumn{1}{c}{$D$}&
	\multicolumn{1}{c}{Layers }&
	\multicolumn{1}{c}{Parameter Amount }&\\
	\midrule
	ConvLSTM & $4\times4$ & 128 &4& 18.87M \\
	TrajGRU & $4\times4$ & 128 &4& 20.78M\\
	ST-ConvLSTM & $4\times4$ & 64 &4&  11.23M \\
	SA-ConvLSTM & $4\times4$ & 64 &4& 9.28M \\
	PredRNN-v2 & $4\times4$ & 128 &4& 42.930M \\
	\midrule
    PTCT & $20\times20$ & 800 &6& 488.48M \\
	\hline
\end{tabular}%
\end{table}%
\section{Influence of Fixed Patch Size under Different Methods}
To investigate the influence of patch size under different patch size, we fix the patch size as $20\times20$ and present results in Table \ref{table:fixed-patch-size}. We also list the parameter amounts of different models. Comapred with ConvLSTM and TrajGRU, PTCT is equipped with slightly more parameters. However, ST-ConvLSTM, SA-ConvLSTM and PredRNN-v2 requires nearly double parameters of PTCT.

 All the experiments are conducted on Radar Echo Guangzhou. Compared with Table 1, ConvLSTM, TrajGRU, ST-ConvLSTM and SA-ConvLSTM improve their performance with large patch size while PredRNN-V2, which reveals the effectiveness of splitting original frames into patches. And our proposed PTCT still remains the best performance, which indicates the importance of Transformer structure proposed in PTCT.

\begin{table}[!h]
\centering
\caption{Performances of different methods under fixed patch size $20\times20$.}
\label{table:fixed-patch-size}%
\begin{tabular}{lccccc}
	\toprule
	\multicolumn{1}{c}{MODEL}&
	\multicolumn{1}{c}{SSIM }&
	\multicolumn{1}{c}{CSI }&
	\multicolumn{1}{c}{POD }&
	\multicolumn{1}{c}{Parameter Amount}&\\
	\midrule
	ConvLSTM & 0.869 & 0.502 & 0.476 & 321.53M\\
	TrajGRU & 0.755 & 0.396 & 0.443 & 405.26M\\
	ST-ConvLSTM & 0.694 & 0.385 & 0.438 & 824.25M\\
	SA-ConvLSTM & 0.711 & 0.362 & 0.431 & 775.78M\\
	PredRNN-v2  & 0.544 & 0.211 & 0.241 & 837.89M\\
	\midrule
    PTCT & \textbf{0.946} & \textbf{0.522} & \textbf{0.582} & 488.48M \\
	\hline
\end{tabular}

\end{table}

\section{More Results on HKO-7}

\begin{figure}[htbp]
	\centering
	
	\includegraphics[width=0.32\linewidth]{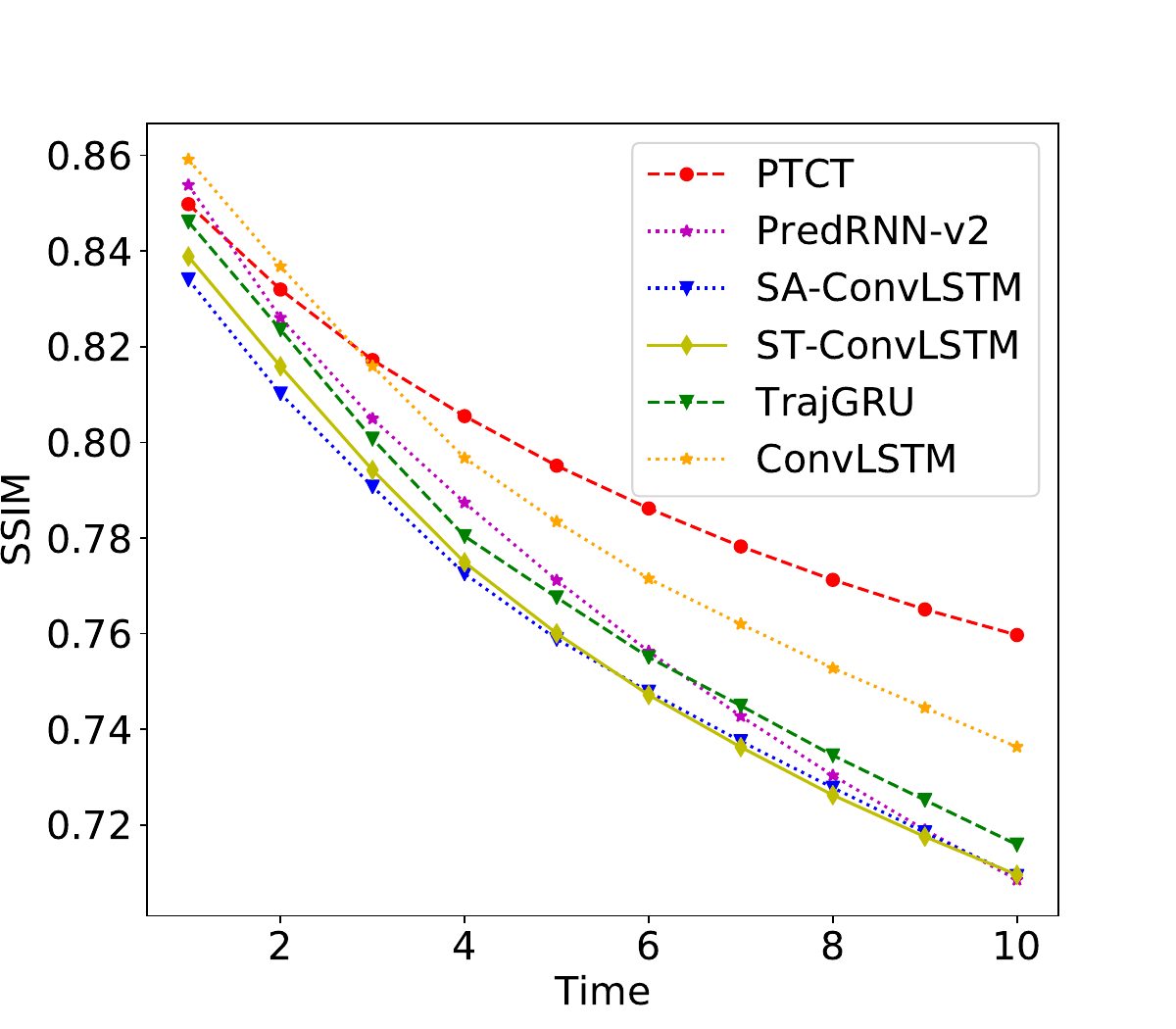}
	\includegraphics[width=0.32\linewidth]{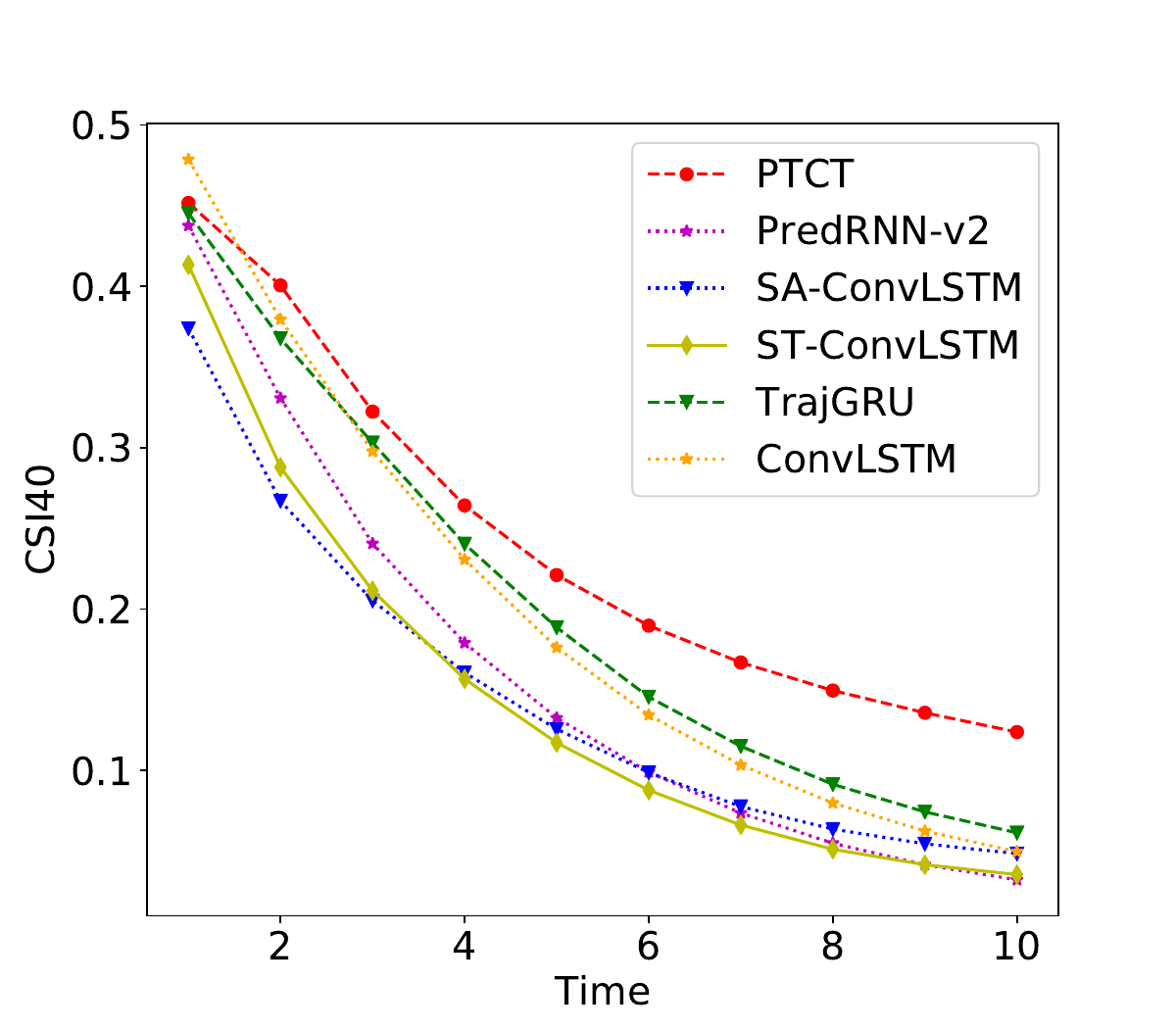}
	\includegraphics[width=0.32\linewidth]{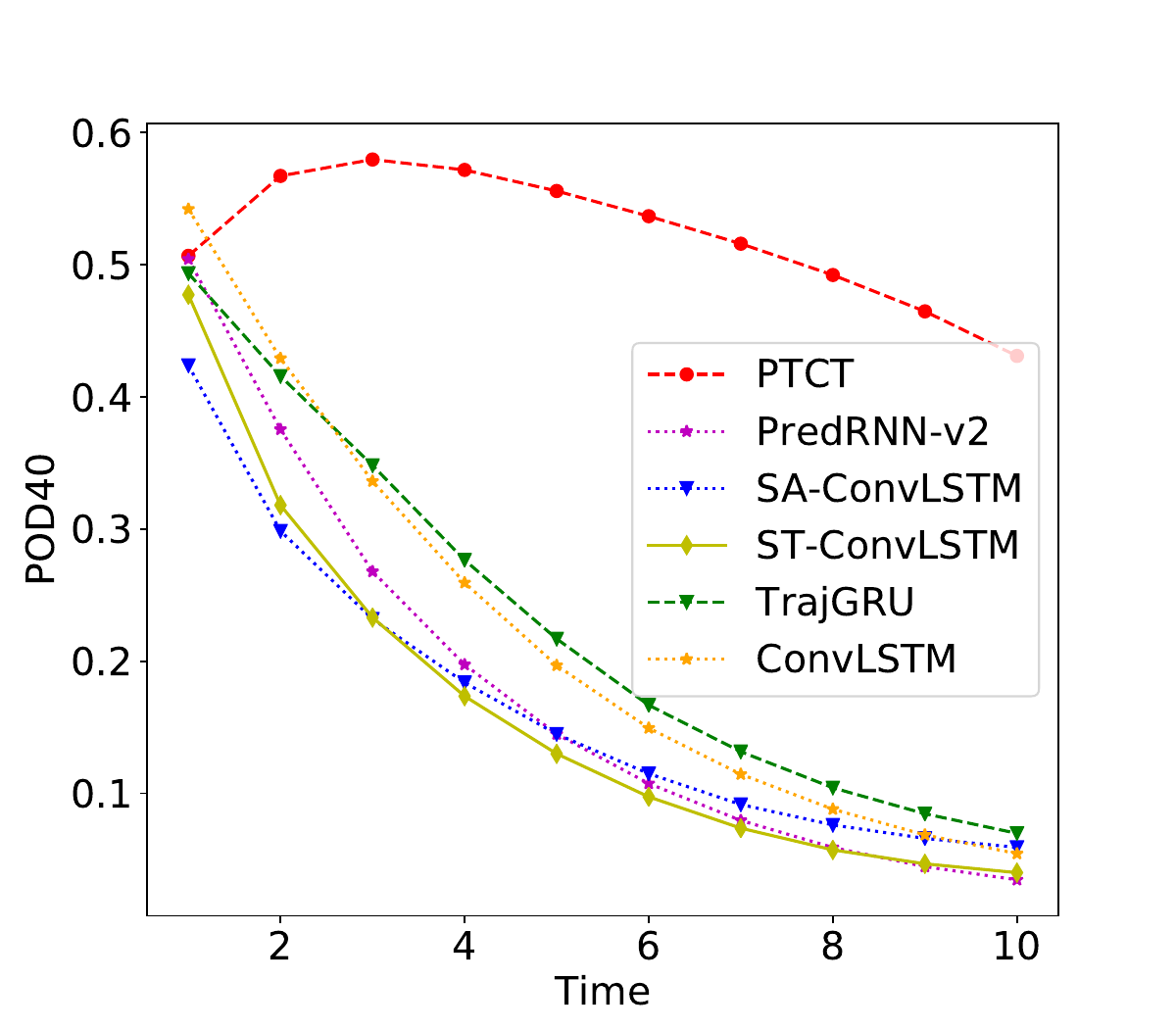}
	\caption{Frame-wise SSIM, CSI and POD on HKO-7.}\label{fig:frame-wise-hko}
\end{figure}

\begin{figure}[htbp]
\centering
\includegraphics[width=0.8\linewidth]{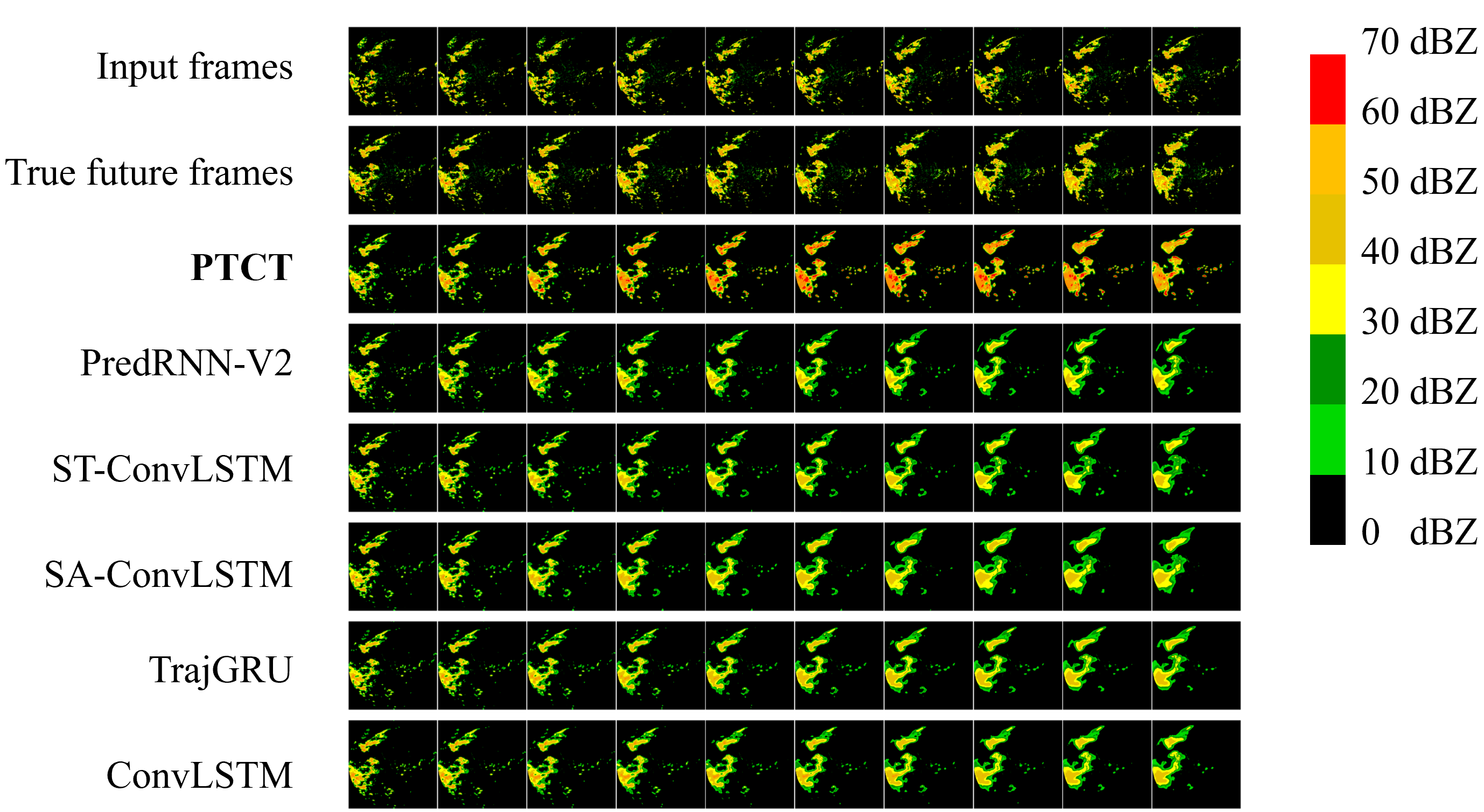}
\caption{Predicted frames on HKO-7.}\label{fig:prediction-hko-seq}
\end{figure}

We plot frame-wise metrics in Figure \ref{fig:frame-wise-hko}, where PTCT performs best in SSIM, CSI and POD almost all the time. To show the effectiveness of PTCT on HKO-7, we plot the predicted frames in Figure \ref{fig:prediction-hko-seq}. Compared with other models, PTCT preserves high sensitivity of strong radar echo, which is critical to predict extreme weather.

\end{document}